\documentclass[a4paper]{svproc}

\newcommand{\figref}[1]{{Fig.~\ref{#1}}}
\newcommand{\tabref}[1]{{Tab.~\ref{#1}}}
\newcommand{\equref}[1]{{Eq.~\ref{#1}}}

\newcommand{\refsec}[1]{Sec. \ref{sec:#1}}
\newcommand{\secref}[1]{\refsec{#1}} 
\newcommand{\refsubsec}[1]{Sec. \ref{subsec:#1}}
\newcommand{\subsecref}[1]{\refsubsec{#1}} 

\newcommand{\argmin}{\mathop{\rm arg~min}\limits}
\usepackage{ulem}

\usepackage[dvipdfmx]{color}

\usepackage[dvipdfmx]{graphicx} 
\usepackage{siunitx}

\newcommand{\upscriptframe}[1]{{}^{\{#1\}}}
\newcommand{\subscriptframe}[1]{_{\{#1\}}}

\usepackage{url}
\usepackage{times}
\usepackage{bm}
\usepackage{amssymb}
\usepackage{amsmath} 

\newif\ifdraft
\draftfalse 

\newcommand{\revise}[1]{\textcolor{\ifdraft red\else black\fi}{#1}}

\begin{document}
\mainmatter              

\title{\LARGE \bf Vectorable Thrust Control for Multimodal Locomotion of Quadruped Robot SPIDAR}
\titlerunning{Vectorable Thrust Control for SPIDAR}  
%
\author{Moju Zhao}
\authorrunning{Moju Zhao} 

\institute{The University of Tokyo, 7-3-1 Hongo, Bunkyo-ku, Tokyo 113-8656, Japan,\\
\email{chou@dragon.t.u-tokyo.ac.jp},\\
\texttt{http://www.dragon.t.u-tokyo.ac.jp/}}

\maketitle              

\begin{abstract}
  In this paper, I present vectorable thrust control for different locomotion modes by a novel quadruped robot, {\bf SPIDAR}, equipped with vectoring rotor in each link.
  First, the robot's unique mechanical design, the dynamics model, and the basic control framework for terrestrial/aerial locomotion are briefly introduced.
  Second, a vectorable thrust control method derived from the basic control framework for aerial locomotion is presented. A key feature of this extended flight control is its ability to avoid interrotor aerodynamics interference under specific joint configuration. Third, another extended thrust control method and a fundamental gait strategy is proposed for special terrestrial locomotion called crawling that requires all legs to be lifted at the same time. Finally, the experimental results of the flight with a complex joint motion and the repeatable crawling motion are explained, which demonstrate the feasibility of the proposed thrust control methods for different locomotion modes.

\keywords{quadruped robot, motion control, multimodal locomotion}
\end{abstract}

\section{Introduction}
\label{sec:introduction}

\begin{figure}[t]
  \begin{center}
    \includegraphics[width=1.0\columnwidth]{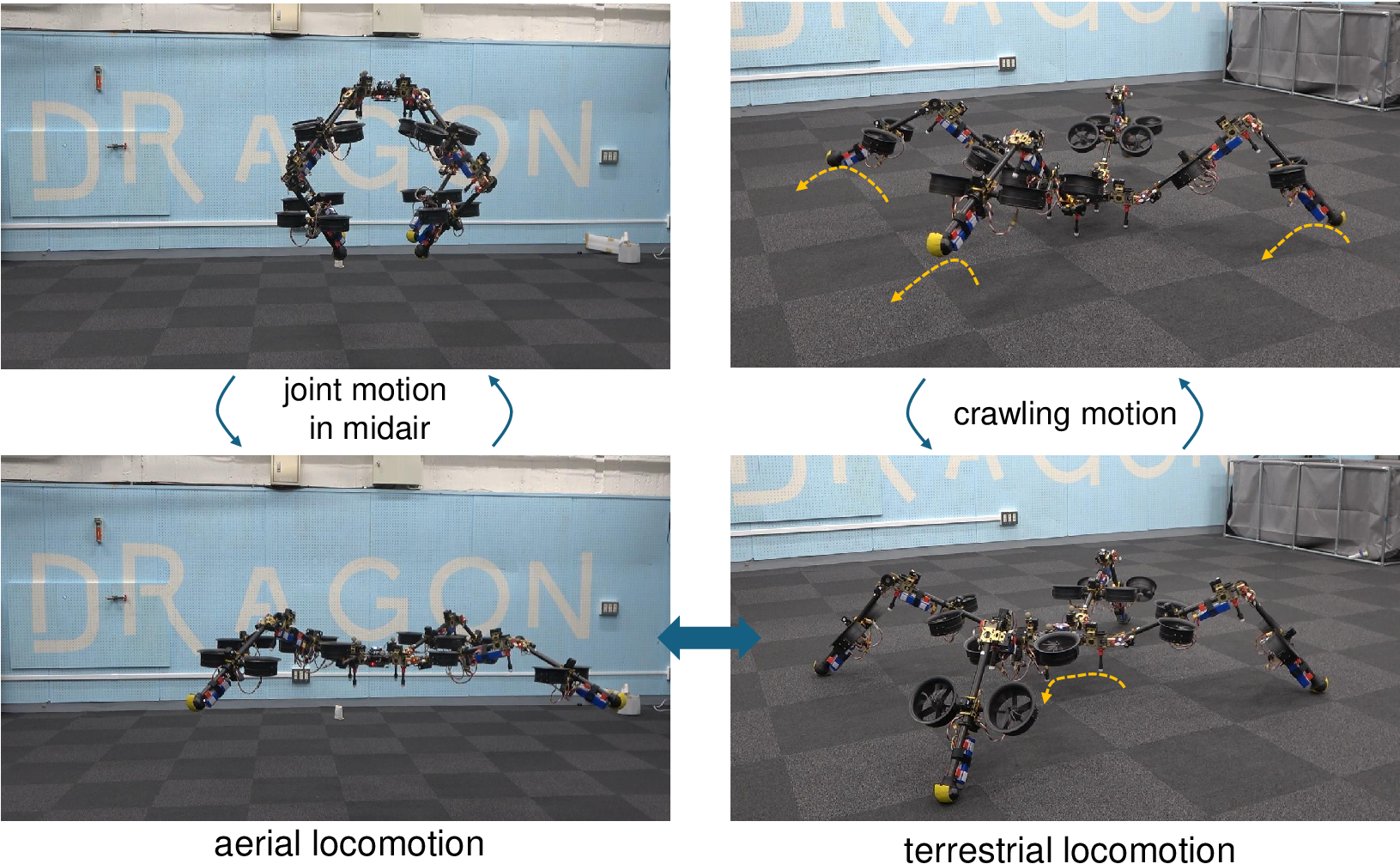}
    \caption{{\bf Aerial-terrestrial quadruped robot SPIDAR}: {\bf (A)} change of the joint configuration in the midair; {\bf (B)} unique crawling motion on the ground by the assistance of vectorable thrust force.}
    \label{figure:intro}
    \vspace{-7mm}
  \end{center}
\end{figure}

During the last decade, robots with multimodal locomotion capability have undergone significant advancements \cite{multimodal-locomotion:MUQA-IROS2013}\cite{multimodal-locomotion:Snake-ACM-R5-ISR2005}\cite{multimodal-locomotion:LoonCopter-JFR2018}.
Among various forms for multimodal locomotion, the legged type \cite{flying-bipedal-robot:LEONARDO-SciRo2021}\cite{flying-bipedal-robot:KXR-RAL2023}\cite{flying-bipedal-robot:Jet-HR2-RAL2022} excels in 1) walking on the unstructured terrains and 2) performing manipulation using limb end-effectors. Then, multilegged (multilimbed) robots are generally more effective due to increased stability in terrestrial locomotion and enhanced freedom in manipulation.
A novel quadruped platform called SPIDAR is proposed in our previous work \cite{aerial-robot:SPIDAR-RAL2023}, which can both walk and fly by using the spherically vectorable rotors distributed in each link unit.
Although a basic thrust control framework is developed in this previous work, it is limited to simple flight motion without the large change in joint angles. For the terrestrial locomotion, a static walk with creeping gait is performed; however, this robot still has the potential to perform other locomotion style for terrestrial movement.
Thus, further investigation into thrust control using vectorable rotors is needed to enable complex midair joint motion and unique terrestrial locomotion, as shown in \figref{figure:intro}.

The link structure of SPIDAR is identical to the aerial robot DRAGON \cite{aerial-robot:DRAGON-RAL2018}, in which the vectorable rotor module is embedded in each link unit. \revise{Unlike the robot platform where the vectoring apparatuses are only equipped at a part of body (e.g., the end of limbs) \cite{flying-bipedal-robot:Jet-HR2-RAL2022}\cite{flying-bipedal-robot:ironcub-RAL2022}, the overlapping between rotors along the vertical direction is a critical issue for SPIDAR}. Such overlapping causes the interrotor aerodynamics interference due to the downwash airflow from the upper rotor, which significantly effects the flight stability. A planning method to avoid the joint configuration that induces rotor overlapping is proposed in \cite{aerial-robot:DRAGON-IROS2018}; whereas \cite{aerial-robot:DRAGON-RAL2020} proposes a realtime compensation control method instead of avoidance. However, for some application such as grasping object, the downwash airflow acting on the target object can significantly effect the grasping stability, which is difficult to solve by the existing methods. Thereby, it is necessary to develop a new solution to prevent the aerointerference between an upper rotor and other downstream segments. Then, in this work, I investigate the boundary of rotor vectoring range, and further consider it as the explicit constraint in an optimization-based control framework.

The branch topology of SPIDAR also provides the ability of terrestrial locomotion such as walking presented in \cite{aerial-robot:SPIDAR-RAL2023}. Given the lightweight design for flight, the compact but weak joint servo motor cannot solely offer the sufficient torque for joint motion. Thus the hybrid thrust/torque control is developed to enable the static walk. However, all rotors are required to generate thrust forces to lift the robot torso even in the idle (standing) mode. Therefore, a more energy-efficient locomotion style should be designed. A unique locomotion style called crawling, which does not require the torso to be lifted all the time, is proposed by \cite{quadruped-crawling:WAREC-1-HUMANOIDS2016}. Another crucial feature of crwaling is that all legs are lifted and moved to the new positions at the same time. Then, the gravity force acting on each leg can significantly effect the balance of the torso (and thus the whole body). Although increasing the weight of torso is a common solution, it conflicts with the lightweight design for the flying ability. Nevertheless, it is possible to use the thrust forces to solve this problem. Thus, in this work I also develop a vectorable thrust control method to compensate the gravity when all legs are lifted during the crawling motion.

\section{Design, Modeling, and Basic Control Framwork}
\label{sec:basic}

\subsection{Design}

\begin{figure}[!b]
  \begin{center}
    \includegraphics[width=1.0\columnwidth]{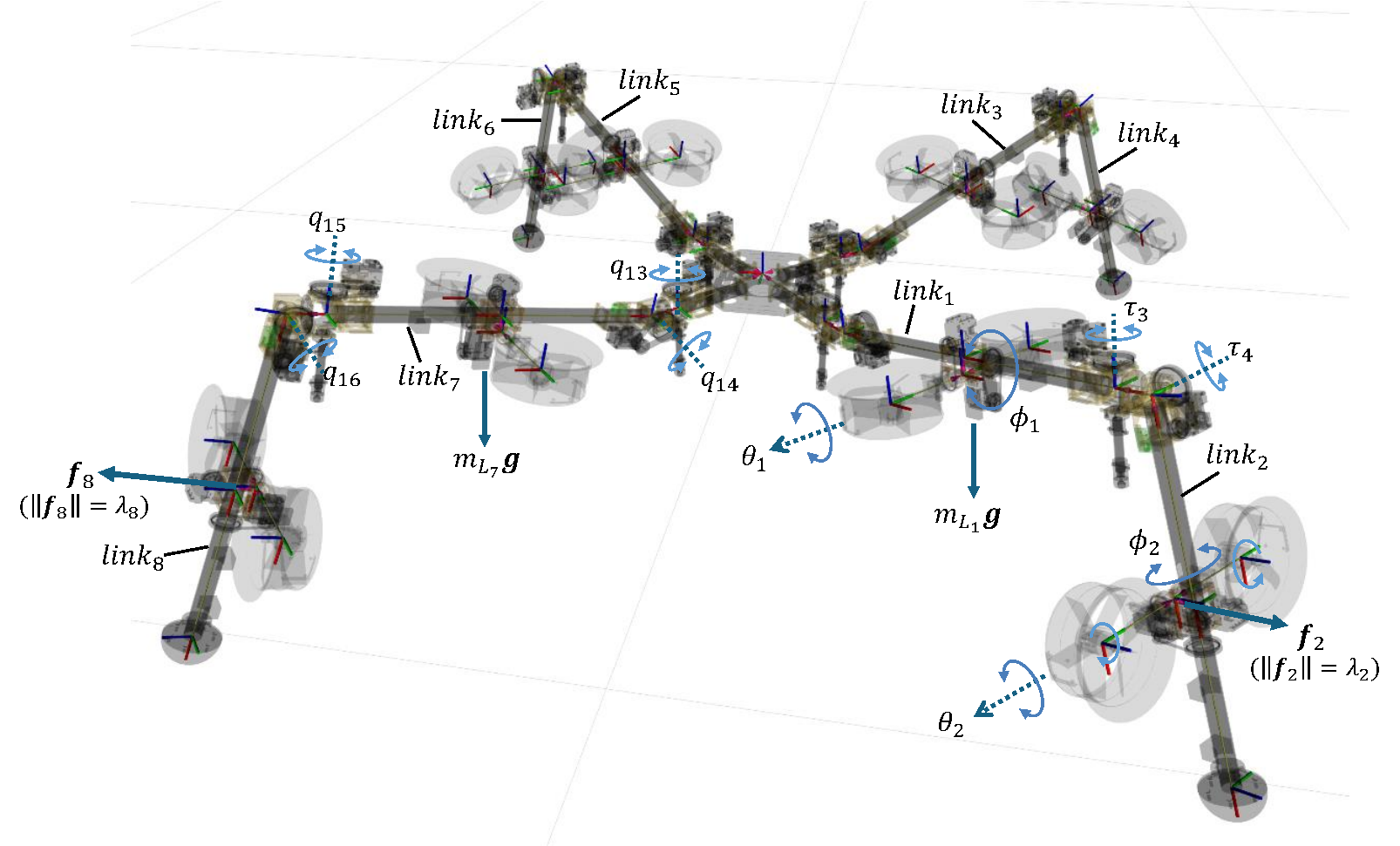}
    \vspace{-5mm}
    \caption{{\bf Kinematics model of the quadruped robot SPIDAR}. There are 4 legs ($l \in \{0, 1, 2, 3\}$) with 8 links. A two-DoF joint module, where the yaw axis comes first followed by the pitch axis, connect neighboring links. The joint servo motor generates joint torque $\tau_i$. Each link contains a spherically vectorable rotor apparatus with two vectoring angles ($\phi, \theta$). Thrust $\lambda_i$ combines a pair to forces from the counter rotating dual rotors.}
    \label{figure:design}
    \vspace{-5mm}
  \end{center}
\end{figure}

\subsubsection{Skeleton Model}
\label{subsec:skeleton_model}

SPIDAR has a point symmetric structure to enable omni-directional movement. Each leg consists of two links that have the same length, and is connected to the center torso. Each joint module has two DoF. Thus, this robot is composed of 8 links with 16 joints as shown in \figref{figure:design}. Given that lightweight design is crucial for aerial locomotion, a compact servo motor is equipped to actuate each joint axis.

\subsubsection{Spherically Vectorable Rotor}
The rotor is required to point arbitrary direction. In other words, it is required to generate a three-dimensional thrust force by each rotor module. Then a spherically vectorable apparatus proposed by \cite{aerial-robot:DRAGON-RAL2018} is equipped in each link as depicted in \figref{figure:design}.
To achieve the spherical vectoring, two rotation axes are necessary. A vectoring axis $\phi_i$ is first introduced around the link rod, which is followed by an orthogonal axis $\theta_i$. Furthermore, a counter-rotating dual-rotor module is applied to counteract the drag moment and gyroscopic moment. Each vectoring axis is also actuated by an individual servo motor. Regarding the thrust force, a combined thrust $\lambda_i$ is introduced for each spherically vectorable rotor module. Eventually, there are three control input ($\phi_i, \theta_i$, and $\lambda_i$) for each vectorable rotor module.

Based on the kinematic model, the three-dimensional thrust force $\bm{f}_i$ w.r.t the frame $\{CoG\}$ is expressed as:
\begin{align}
  \label{eq:thrust_model}
  \bm{f}_{i} &=  \lambda_i  \bm{u}_{i}(\phi_i, \theta_i),
\end{align}
where $\bm{u}_{i}$ denotes the unit vector for the spherically vectorable mechanism that is effected by two vectoring angles $\phi_i$ and $\theta_i$.

Then the total wrench in the frame $\{CoG\}$ can be given by
\begin{align}
  \label{eq:total_wrench}
\begin{bmatrix} \bm{f}_{\lambda} \\ \bm{\tau}_{\lambda} \\ \end{bmatrix}
= \begin{bmatrix}
  \bm{u}_1  & \cdots & \bm{u}_{N_{\mathrm{r}}} \\
  \bm{p}_{1} \times \bm{u}_1& \cdots & \bm{p}_{N_{\mathrm{r}}} \times \bm{u}_{N_{\mathrm{r}}}
\\ \end{bmatrix}
  \bm{\lambda},
\end{align}
where $\bm{p}_{i}$ is the rotor position that is influenced by the joint angles $\bm{q}$.

\subsection{Modeling}
\label{subsec:modeling}


The joint motion is assumed to be relatively slow, then the quasi-static assumption ($\dot{\bm{q}} \approx \bm{0}; \ddot{\bm{q}} \approx \bm{0}$) can be applied. Then the whole-body dynamics can be approximated as follows::
\begin{align}
  \label{eq:approx_translational_dynamics}
&m_{\Sigma} \ddot{\bm r}(\bm{q}) = \bm{R} \bm{f}_{\lambda}
- m_{\Sigma} {\bm g}, \\
\label{eq:approx_rotational_dynamics}
&\bm{I}_{\Sigma}(\bm{q})\dot{\bm \omega} + {\bm \omega} \times \bm{I}_{\Sigma}(\bm{q}) {\bm \omega} = \bm{\tau}_{\lambda}, \\
\label{eq:approx_joint_dynamics}
& \bm{0} = \displaystyle \bm{\tau}_{q}
  +  \sum^{N_r}_{i=1} \bm{J}_{\mathrm{r}_i}^{\mathsf{T}} \bm{f}_i
  + \sum^{N_s}_{i=1} \bm{J}_{\mathrm{s}_i}^{\mathsf{T}} m_{\mathrm{s}_i}\bm{g},
\end{align}
where ${\bm r}(\bm{q})$ is the position of the frame $\{CoG\}$ w.r.t the frame $\{W\}$.
$ \bm{R} $ is the orientation of the frame $\{CoG\}$ w.r.t. the frame $\{W\}$.
$m_{\Sigma}$ is the total mass.
${\bm \omega}$ is the angular velocity of the frame $\{CoG\}$  w.r.t the frame of $\{CoG\}$. $\bm{I}_{\Sigma} (\bm{q})$ is the total inertia tensor that is also influenced by the joint angles. $\bm{J}_{{\mathrm{*}}_i}\in{\mathcal R}^{3 \times N_J}$ is the Jacobian matrix for the frame of the $i$-th rotor ($* \rightarrow \mathrm{r}$) and the $i$-th segment's CoG ($* \rightarrow \mathrm{s}$), respectively. $\bm{\tau}_q \in \mathcal{R}^{N_J}$ is the vector of joint torque.

\subsection{Basic Control Framework}

\subsubsection{Centroidal Motion Control}

For the approximated centroidal dynamics of \eqref{eq:approx_translational_dynamics} and \eqref{eq:approx_rotational_dynamics}, the position feedback control based on an ordinary PID control is given by
\begin{align}
  \label{eq:pid_pos}
  {\bm f}_{\lambda}^d =  m_{\Sigma} \bm{R}^{\mathsf{T}} \left(k_{f, p} \bm{e}_{\bm{r}} + k_{f, i} \int \bm{e}_{\bm{r}} + k_{f, d} \dot{\bm{e}}_{\bm{r}} + \bm{g}\right),
\end{align}
where $\bm{e}_{\bm{r}} = {\bm r}^d - {\bm r}$, and $k_{f, \ast}$ are the PID gains.

The attitude control follows the part of the control method proposed by \cite{aerial-robot:SE3-Control-CDC2010}:
\begin{align}
  \label{eq:pid_rot}
  {\bm \tau}^d_{\lambda} = \bm{I}_{\Sigma} \left(k_{\tau, p} \bm{e}_{R} + k_{\tau, i} \int \bm{e}_{R} + k_{\tau, d} \bm{e}_{\bm{\omega}}\right) + {\bm \omega} \times \bm{I}_{\Sigma} {\bm \omega}, \\
  \bm{e}_{R} = \frac{1}{2}\left[\bm{R}^{\mathsf{T}}\bm{R}^d - \bm{R}^{d\mathsf{T}}\bm{R}\right]^{\vee},
  \bm{e}_{\bm{\omega}} = \bm{R}^{\mathsf{T}}\bm{R}^d\bm{\omega}^d - \bm{\omega}, \nonumber
\end{align}
where $\left[\star\right]^{\vee}$ is the inverse of a skew map.

Then, the desired wrench w.r.t the frame $\{CoG\}$ can be summarized as follows:
\begin{align}
  \label{eq:desired_wrench}
  \bm{\mathrm{w}} ^d = \begin{bmatrix} \bm{f}_{\lambda} ^d && \bm{\tau}_{\lambda} ^d \end{bmatrix}^{\mathsf{T}}.
\end{align}

\begin{figure}[t]
  \begin{center}
    \includegraphics[width=0.8\columnwidth]{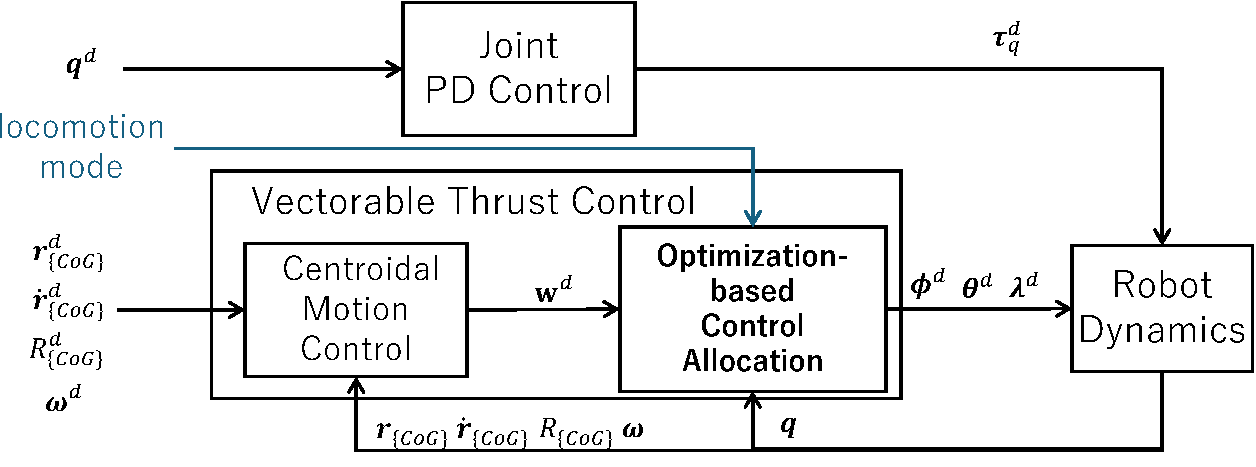}
    \vspace{-3mm}
    \caption{{\bf Basic control framework for terrestrial/aerial locomotion}. The vectorable thrust control and the joint control are performed independently. The cost function and the constraints of the optimization-based control allocation would slightly modified according to the locomotion mode (aerial locomotion in \secref{flight} and terrestrial locomotion in \secref{crawl}).}
    \vspace{-5mm}
    \label{figure:control}
  \end{center}
\end{figure}

\subsubsection{Control Allocation}
\label{subsec:control_allocation}

As shown in \figref{figure:control}, the goal of thrust control is to obtain the control input ($\bm{\phi}^d, \bm{\theta}^d$, and $\bm{\lambda}^d$) from the desired wrench ${\bm{\mathrm{w}}}^d$. Meanwhile, it is also important to suppress the rotor output and the joint load from the aspect of the energy consumption. Then, an optimization problem is introduced.
Instead of using $\bm{\phi}$ and $\bm{\theta}$ which induce the nonlinearity in the optimization problem, the linear force $\bm{f}_{i}$ is introduced \revise{as proposed in \cite{aerial-robot:DRAGON-IJRR2022}}. Then the optimization problem can be written as follows:
\begin{align}
  \label{eq:rough_allocation_cost}
&   \displaystyle \min_{\bm{f}_{i}, \bm{\tau}_q}  \hspace{3mm} w_1  {\displaystyle \sum_{i = 1}^{N_{\mathrm{r}}}} \| \bm{f}_{i} \|^2 + w_2 \|\bm{\tau}_q\|^2,\\
  \label{eq:wrench_allocation_constraint}
&   s.t.   \hspace{3mm}
  \bm{\mathrm{w}} ^d  = {\displaystyle \sum_{i = 1}^{N_{\mathrm{r}}}} \bm{Q}_{i} \bm{f}_{i},
  \hspace{3mm}
  \bm{Q}_{i} =
  \begin{bmatrix}
    \bm{I}_{3\times 3} \\
    \left[\bm{p}_i \times \right]
  \end{bmatrix} ,\\
  \label{eq:quasi_static_constraint}
  &  \hspace{8mm} \displaystyle \bm{\tau}_{q} =
  -  \sum^{N_{\mathrm{r}}}_{i=1} \bm{J}_{\mathrm{r}_i}^{\mathsf{T}} \bm{f}_i
  - \sum^{N_s}_{i=1} \bm{J}_{\mathrm{s}_i}^{\mathsf{T}} m_{\mathrm{s}_i}\bm{g}, \\
  \label{eq:thrust_constraint}
  & \hspace{8mm} -\bar{\lambda} \leq f_{i,j} \leq \bar{\lambda},  \\
  \label{eq:joint_constraint}
  & \hspace{8mm} - \bar{\tau}_q \leq \tau_{q_i} \leq \bar{\tau}_q,
\end{align}
where $w_1$ and $w_2$  in \equref{eq:rough_allocation_cost} are the weights for the cost of rotor thrust and joint torque, respectively.
\equref{eq:wrench_allocation_constraint} is obtained by substituting \eqref{eq:desired_wrench} into \equref{eq:total_wrench}.
\equref{eq:quasi_static_constraint} denotes the equilibrium between the joint torque $\bm{\tau}_q$, the thrust force $\bm{f}_i$, and the segment gravity $m_{\mathrm{s}_i}\bm{g}$.
\eqref{eq:thrust_constraint} denotes the constraint of thrust force. Instead of the nonlinear bound ($\| \bm{f}_{i} \| \leq \bar{\lambda}$), a cubic bound is introduced.
\equref{eq:joint_constraint} denotes the bound for the joint torque.

Once the optimized thrust force $\bm{f}_{i}$ is calculated, the true control input for the spherically vector rotor apparatus can be obtained as follows:
\begin{align}
  \label{eq:pseudo_inverse_desired_thrust_force}
  &\lambda_i = \| \bm{f}_{i} \| ,\\
  \label{eq:pseudo_inverse_desired_vectoring_phy}
  &\phi_i = \text{tan}^{-1}\left(\frac{-f_{i,y}}{f_{i,z}}\right),\\
  \label{eq:pseudo_inverse_desired_vectoring_theta}
  &\theta_{i} = \text{tan}^{-1}\left(\frac{f_{i,x}}{-f_{i,y} sin(\phi_{i}) + f_{i,z} cos(\phi_{i})}\right).
\end{align}

\subsubsection{Joint Control}
The proposed optimization problem of \equref{eq:rough_allocation_cost} can also provides the joint torque that however only satisfies the quasi-static assumption for joint motion.
Hence, it is necessary to apply a feed-back control to track the desired position for joints. Therefore, a simple PD control for joint position is introduced for each joint:
\begin{align}
  \label{eq:joint_pd_control}
\tau_{q_i}^d  =  k_{\mathrm{j},p}(q_i^d - q_i) - k_{\mathrm{j},d} \dot{q}_i.
\end{align}

\section{Flight Motion}
\label{sec:flight}

\subsection{Interrotor Aerointerference}

\begin{figure}[t]
  \begin{center}
    \includegraphics[width=1.0\columnwidth]{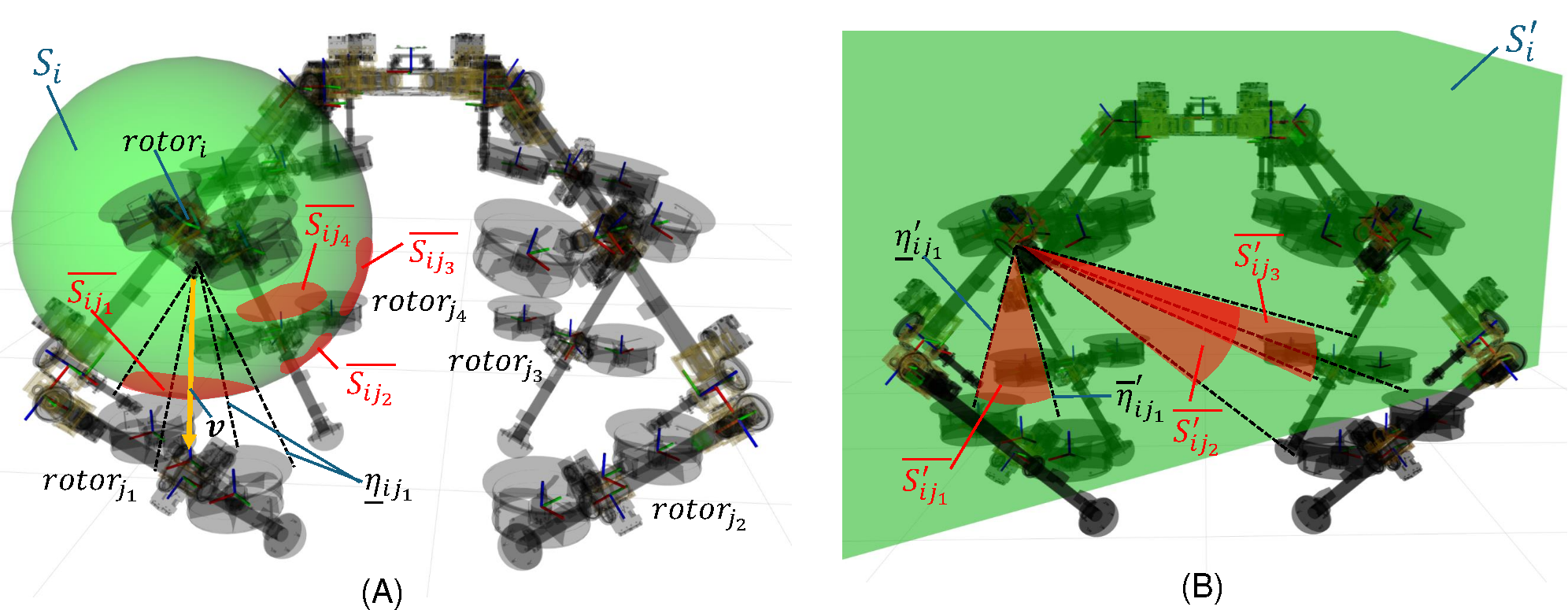}
    \vspace{-7mm}
    \caption{{\bf Vectoring range of $rotor_i$, where green area denotes the valid range, and the red area denotes the invalid range that causes the aerointerference with other rotors.} {\bf (A)} spherical vectoring range expressed in three-dimensional manner; {\bf (B)} planar vectoring range $S^{'}_i$ expressed in two-dimensional manner.}
    \label{figure:interference}
    \vspace{-5mm}
  \end{center}
\end{figure}

In order to avoid the aerointerference caused by downwash from upper rotors effecting downstream segments, the rotor vectoring range is constrained.
The valid vectoring range for a rotor module can be described as follows:
\begin{align}
  S_i &:=  \bigcap S_{ij}, \\
  S_{ij} &: = \{\theta_i, \phi_i | \eta_i(\theta_i, \phi_i) > \underline{\eta}_{ij}(\theta_i, \phi_i) \},
\end{align}
where $\eta_i(\theta_i, \phi_i)$ is the unit vector of the rotor direction, and $\underline{\eta}_{ij}(\theta_i, \phi_i)$ can be considered as the boundary of the invalid area which is depicted as the green area in \figref{figure:interference}(A).

\begin{figure}[h]
  \begin{center}
    \includegraphics[width=1.0\columnwidth]{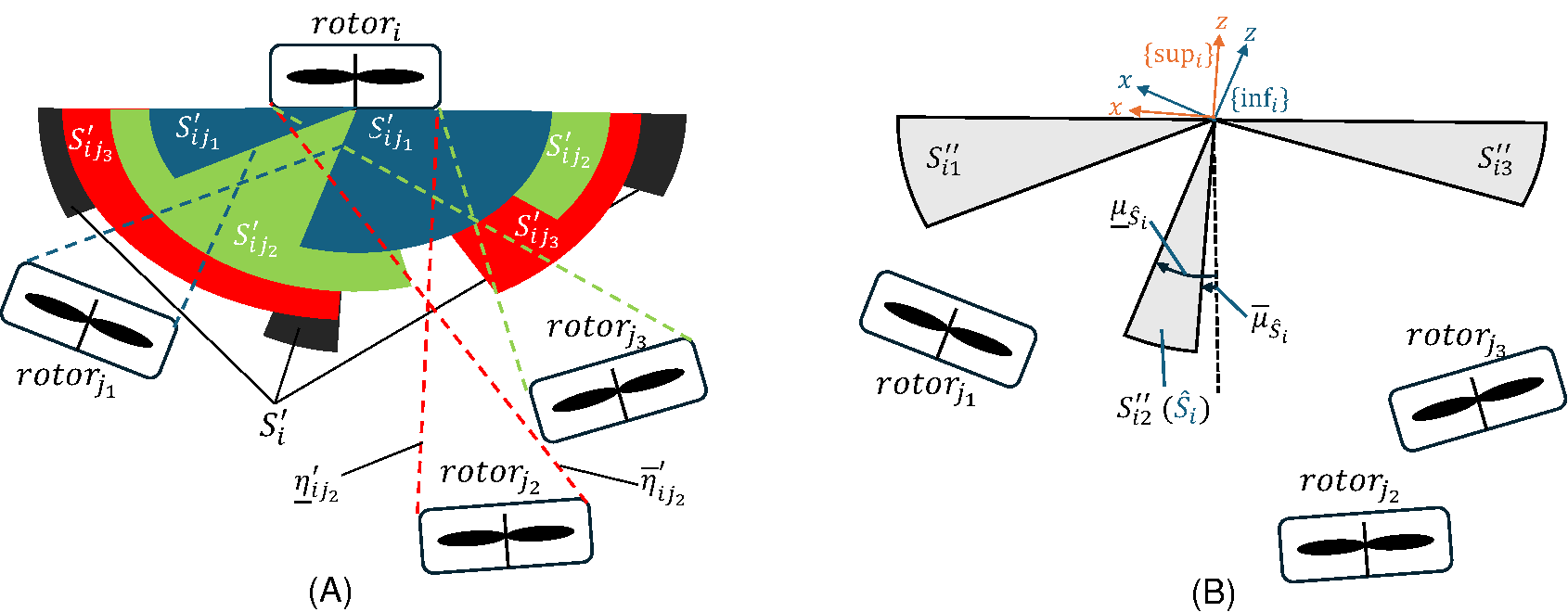}
    \caption{{\bf Simplified aerointerference model by fixing the vectoring angle $\phi_i$ at $\hat{\phi}_i$.} {\bf (A)} the description of $S^{'}_{ij}$; {\bf (B)} the description of $S^{''}_{ij}$.}
    \label{figure:simple_interference}
  \end{center}
\end{figure}

Although $\underline{\eta}_{ij}(\theta_i, \phi_i)$ can be theoretically computed, its nonlinear boundary poses challenges in the control framework described in \subsecref{control_allocation}. Thus, as shown in \figref{figure:interference}(B), it is desired to find a principal axis to decrease the spherical range to a planer range, which can be written by
\begin{align}
  S^{'}_i &:=  \bigcap S^{'}_{ij}, \\
  S^{'}_{ij} &:= \{\psi_i | \psi_i < \underline{\eta}^{'}_{ij}, \bar{\eta}^{'}_{ij} < \psi_i \}.
\end{align}
\revise{To find a valid principal vectoring angle $\psi_i$, the near-hovering state, where rotors tilt around their upward direction within a certain range, is introduced, and the pair of nominal vectoring angles ($\hat{\phi}_{i}, \hat{\theta}_{i}$) that makes the rotor upward can be found by solving $\bm{u}(\hat{\theta}_{i}, \hat{\phi}_{i}) =  \begin{bmatrix} 0 && 0 && 1 \end{bmatrix}^{\mathsf{T}}$.
If the first vectoring angle ${\phi}_{i}$ is fixed at $\hat{\phi}_{i}$, the second vectoring angle $\theta_i$ will be a candidate of $\psi_i$ that can minimize the forbidden areas like $\overline{S^{'}_{ij_1}}$ in \figref{figure:simple_interference}(B) owing to its vectoring axis parallel to the long side of the rotor module.}

To quantify the relative position between two rotors, a vector $\bm{v}$ is further introduced to connect a pair of rotors as shown in \figref{figure:interference}(A). Then the inclination along the first vectoring angle ${\phi}_{i}$ can be roughly given by
\begin{align}
  \label{eq:approx_roll_angle}
  \alpha = tan^{-1}(\frac{v_y}{v_z}).
\end{align}
  If $\alpha$ is larger than a certain thresh $\underline{\alpha}$, the aerointerference between this pair of rotors can be ignored in near-hovering state (i.e., $\phi_i \approx \hat{\phi}_i$).

As depicted in \figref{figure:simple_interference}(A), the valid range of the $i$-th rotor can be eventually given by
\begin{align}
  S^{'}_{ij} := \{\theta_i | \theta_i < \underline{\eta}^{'}_{ij},  \bar{\eta}^{'}_{ij} < \theta_i \}, \hspace{3mm} j \in \{ \mathbb{N} |\alpha_{ij} < \underline{\alpha} \},
\end{align}
where $\underline{\eta}^{'}_{ij}$ and $\bar{\eta}^{'}_{ij}$ are easy to calculate from the relative position between two rotors.
\revise{Given that $\alpha$ is determined solely by quasi-static joint angles $\bm{q}$ (where $\dot{\bm{q}} \approx \bm{0}; \ddot{\bm{q}} \approx \bm{0}$) , it remains constant during the control allocation as described in \eqref{eq:rough_allocation_cost}$\sim$\eqref{eq:joint_constraint}.}

\subsection{Vectored Thrust Control}
\label{subsec:vectored_thrust_control}

Once $S^{'}_{i}$ is obtained, we can rewrite it as a combination of several isolated subsets $S^{''}_{ik}$:
\begin{align}
  S^{'}_i &:=  \bigcup S^{''}_{ik}, \\
  S^{''}_{ik} &:= \{\theta_i | \underline{\mu}_{ik} < \theta_i < \bar{\mu}_{ik}\},
\end{align}
where $S^{''}_{ik_1} \cap S^{''}_{ik2} = \varnothing$.

In order to avoid the sudden jumping between separate subsets as shown in \figref{figure:simple_interference}(B), the vectoring range of $\theta_i$ is restricted within one subset.
The best subset can be considered as the most energy-efficient one.
Given that the majority of the thrust force is used to counteract the gravity, the rotor inclination from the vertical direction corresponds to the energy efficiency (i.e., $cos (\theta_i)$). Therefore, the best subset  can be given by
\begin{align}
  \hat{S}_{i} := \argmin_{S^{''}_{ik}} (\min_{\theta \in S^{''}_{ik}}(\|\theta\|)).
\end{align}

Then, the vectoring angle bound for $\theta_i$ can be obtained by:
\begin{align}
  \label{eq:theta_bound}
  \underline{\mu}_{\hat{S}_{i}} < \theta_i < \bar{\mu}_{\hat{S}_{i}}.
\end{align}

However, this linear constraint for vectoring angle cannot directly substitute to the control framework of \eqref{eq:rough_allocation_cost} $\sim$ \eqref{eq:joint_constraint}.
Therefore, it is necessary to convert this constraint into the three-dimensional force $\bm{f}_{i}$. Here, we can introduce two new coordinates $\{\text{inf}_i\}$ and  $\{\text{sup}_i\}$ that are align with the bounds respectively as depicted in \figref{figure:simple_interference}(B). Then the coordinate transformation for the three-dimensional force can be written as follows:
\begin{align}
  \upscriptframe[0.6]{\text{inf}_i}\bm{f}_{i} = \upscriptframe[0.6]{\text{inf}_i}\bm{R}(\underline{\mu}_{\hat{S}_{i}}) \bm{f}_{i}, \\
  \upscriptframe[0.6]{\text{sup}_i}\bm{f}_{i} = \upscriptframe[0.6]{\text{sup}_i}\bm{R}(\bar{\mu}_{\hat{S}_{i}}) \bm{f}_{i}.
\end{align}.

Then, following linear constraints regarding the force can be obtained:
\begin{align}
  \label{eq:inf_linear_bound}
  \upscriptframe[0.6]{\text{inf}_i}f_{i, x} \leq 0, \\
  \label{eq:sup_linear_bound}
  \upscriptframe[0.6]{\text{sup}_i}f_{i, x} \geq 0.
\end{align}

Besides, since $\phi_i$ is fixed at $\hat{\phi}_i$, the following equal constraints should be also considered:
\begin{align}
  \label{eq:y_linear_bound}
  f_{i,y} = 0.
\end{align}

Again, only the rotor that has the potential aerointerference with other down stream segments should be restricted. Hence, a positive threshold $\underline{\theta}$ is introduced to extract the target rotors. Then the restricted rotor subset $\mathcal{I}$ can be written as:
\begin{align}
  \mathcal{I} = \{i | \min (-\underline{\mu}_{\hat{S}_{i}}, \bar{\mu}_{\hat{S}_{i}})  <  \underline{\theta} \}.
\end{align}

Finally, the optimization problem for control allocation will be modified as follows:
\begin{align}
   \displaystyle \min & \hspace{1mm}  \eqref{eq:rough_allocation_cost}, \nonumber \\
      s.t. & \hspace{1mm} \eqref{eq:wrench_allocation_constraint} \sim \eqref{eq:joint_constraint} \hspace{2mm} \text{for} \hspace{1mm} ^{\forall}i, \nonumber \\
   &  \hspace{1mm} \eqref{eq:inf_linear_bound} \sim \eqref{eq:y_linear_bound} \hspace{2mm} \text{for} \hspace{1mm} i \in \mathcal{I}. \nonumber
\end{align}
\revise{The additional constraints \eqref{eq:inf_linear_bound} $\sim$ \eqref{eq:y_linear_bound} guarantee the convexity of optimization problem, since they are linear and the boundaries of these constraints ($\underline{\mu}_{\hat{S}_{i}}$ and $\bar{\mu}_{\hat{S}_{i}}$) which are calculated from the quasi-static joint angle are constant during optimization.}

\section{Crawling Motion}
\label{sec:crawl}

\subsection{Thrust Vectoring for Lifting Legs}
\label{subsec:crawling_control}
\begin{figure}[h]
  \begin{center}
    \includegraphics[width=1.0\columnwidth]{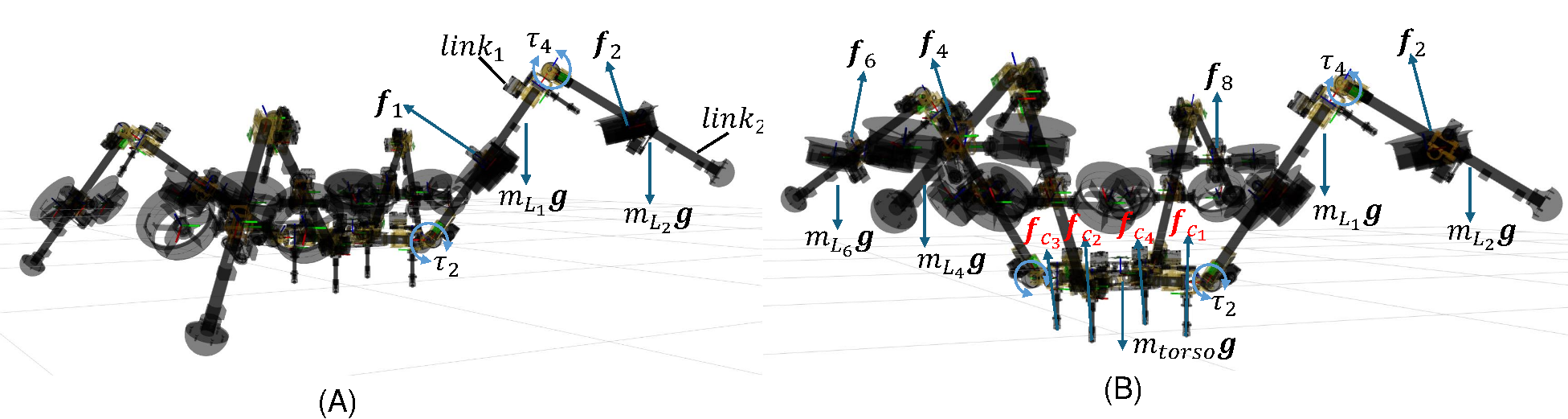}
    \vspace{-7mm}
    \caption{{\bf Lifting legs while the torso is touching to the ground}: {\bf (A)} case of only lifting one leg; {\bf (B)} case of lifting all legs, and the torso mass $m_{torso}$ and the contact forces $\bm{f}_{c}$ acting on the torso should be considered.}
    \label{figure:leg_lift}
  \end{center}
\end{figure}

Crawling locomotion requires simultaneously lifting all legs. The joint motion is assumed to be sufficiently slow, and thus the motion is quasi-static. Then, the key challenge is to balance the lifted legs against gravity using joint torque and the vectorable thrust force, as shown in \figref{figure:leg_lift}(A).

To lift a single leg, it is only necessary to consdier that balance of the target leg. Then a simplified optimization framework can be given by
\begin{align}
   \displaystyle \min & \hspace{1mm}  \eqref{eq:rough_allocation_cost}, \nonumber \\
      s.t. & \hspace{1mm} \eqref{eq:quasi_static_constraint} \sim \eqref{eq:joint_constraint} \hspace{2mm} \text{for} \hspace{1mm} i \in \mathcal{L}, \nonumber
\end{align}
where $\mathcal{L}$ denotes the set of the rotors that are included in the lifting leg.

However, if all legs are lifted at the same time, it is also required to consider the balance of the torso (that is the whole body) as shown in \figref{figure:leg_lift}(B). The total wrench w.r.t. in the baselink frame $\{B\}$ can be given by:
\begin{align}
  \label{eq:torso_balance}
&  \bm{\mathrm{w}} = \sum^{N_c}_{i=1} \hat{\bm{J}}_{\mathrm{c}_i}^{\mathsf{T}} \bm{f}_{c_i}
  +  \sum^{N_r}_{i=1} \hat{\bm{J}}_{\mathrm{r}_i}^{\mathsf{T}}\bm{f}_i
  + \sum^{N_s}_{i=1} \hat{\bm{J}}_{\mathrm{s}_i}^{\mathsf{T}} m_{\mathrm{s}_i}\bm{g}, \\
&    \hat{\bm{J}}_{*} =
    \begin{bmatrix}
      \bm{I}_{3\times 3} \\
    \left[\upscriptframe{B}\bm{p}_{*} \times \right]
    \end{bmatrix} \upscriptframe{B} \bm{R}\subscriptframe[0.6]{*}, \nonumber
\end{align}
where $\bm{f}_{c_i}$ denotes the contact force acting on the torso. In the case of SPIDAR, there are four contact points (i.e., $N_c = 4$). It is notable that if the torso is static, the total wrench $\bm{\mathrm{w}}$ will be $\bm{0}$.

Given that $\bm{f}_{c_i}$ is difficult to obtain without the force sensor, following wrench related to the contact forces is introduced:
\begin{align}
  \label{eq:torso_wrench}
  \bm{\mathrm{w}}_{c} = \sum^{N_c}_{i=1} \hat{\bm{J}}_{\mathrm{c}_i}^{\mathsf{T}} \bm{f}_{c_i}.
\end{align}
Then, the range for this contact wrench can be given by
\begin{align}
  \label{eq:torso_wrench_range}
&   \underline{\bm{\mathrm{w}}} \leq \bm{\mathrm{w}}_{c} \leq  \bar{\bm{\mathrm{w}}}, \\
&   \underline{\bm{\mathrm{w}}} = \begin{bmatrix} -\bar{f}_{xy} & -\bar{f}_{xy} & 0 & -\bar{\tau}_{xy} & -\bar{\tau}_{xy} & -\bar{\tau}_{z} \end{bmatrix}^{\mathsf{T}}, \nonumber \\
&   \bar{\bm{\mathrm{w}}} = \begin{bmatrix} \bar{f}_{xy} & \bar{f}_{xy} & \infty & \bar{\tau}_{xy} & \bar{\tau}_{xy} & \bar{\tau}_{z} \end{bmatrix}^{\mathsf{T}}, \nonumber
\end{align}
where the positive threshold of $\bar{f}_{xy}$  and $\bar{\tau}_{z}$ are mainly related to the friction; whereas $\bar{\tau}_{xy}$ corresponds to the torso mass and the support polygon.

Substitute \eqref{eq:torso_balance} and \eqref{eq:torso_wrench} into \eqref{eq:torso_wrench_range} with $\bm{\mathrm{w}} = \bm{0}$, we can obtain an inequality constraint for thrust force:
\begin{align}
  \label{eq:torso_balance_constraint}
   \underline{\bm{\mathrm{w}}} \leq
  -  \sum^{N_r}_{i=1} \hat{\bm{J}}_{\mathrm{r}_i}^{\mathsf{T}}\bm{f}_i
  - \sum^{N_s}_{i=1} \hat{\bm{J}}_{\mathrm{s}_i}^{\mathsf{T}} m_{\mathrm{s}_i}\bm{g}
  \leq \bar{\bm{\mathrm{w}}}.
\end{align}

Finally, the control allocation for balancing the lifting legs can be summarized as:
\begin{align}
   \displaystyle \min & \hspace{1mm}  \eqref{eq:rough_allocation_cost}, \nonumber \\
      s.t. & \hspace{1mm} \eqref{eq:quasi_static_constraint} \sim \eqref{eq:joint_constraint}, \eqref{eq:torso_balance_constraint}, \nonumber
\end{align}
where \eqref{eq:wrench_allocation_constraint}, which is the constraint regarding the centroidal motion control, is ignored.

The sum of the thrust force $\sum^{N_r}_{i=1} \|\bm{f}_i\|$ can be much smaller that the total weight of the robot $\sum^{N_s}_{i=1} m_{\mathrm{s}_i}\bm{g}$, because the third component in the upper limit of contact wrench  $\bar{\bm{\mathrm{w}}}$ is $\infty$. This indicates that the contact force can offer additional upward force to hold the torso (that is the whole body) against gravity. 

\subsection{Crawling Gait}
\begin{figure}[h]
  \begin{center}
    \includegraphics[width=1.0\columnwidth]{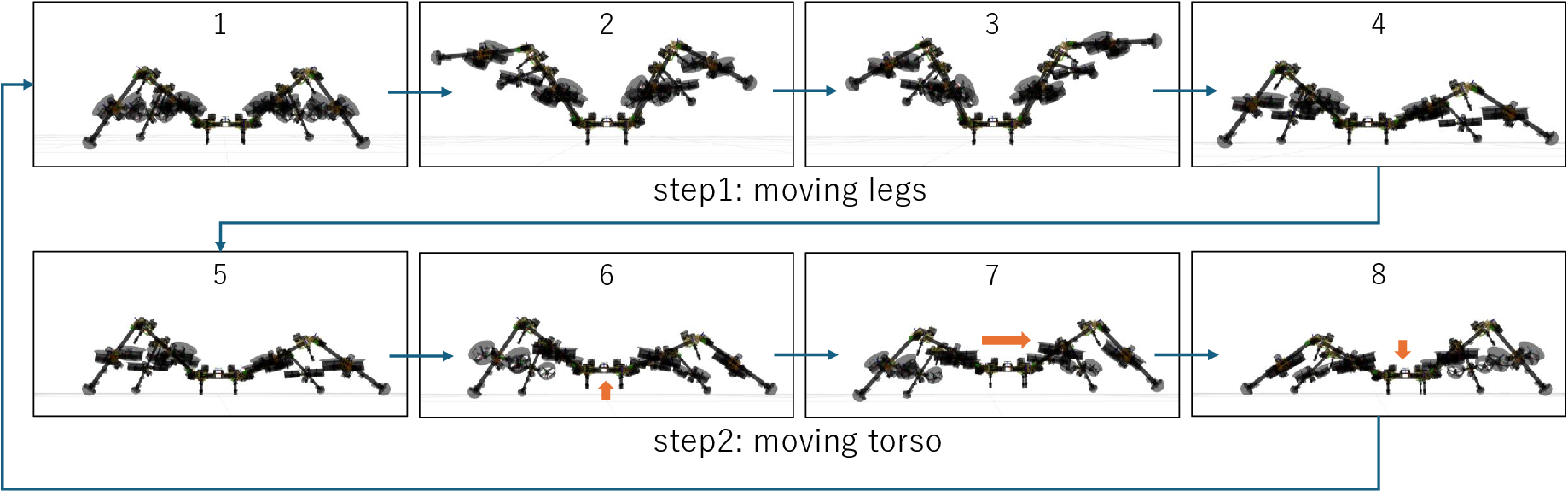}
    \vspace{-5mm}
    \caption{{\bf Crawling gait of the unique quadruped robot SPIDAR}. Two steps in one gait cycle: 1) moving all legs; 2) moving the torso.}
    \label{figure:crawl_gait}
    \vspace{-5mm}
  \end{center}
\end{figure}

As depicted in \figref{figure:crawl_gait}, a single gait cycle can be separated into two steps. The first step is to move legs, which is followed by the second step moving the torso. For a linear movement, the stride length of all feet is set equal to the moving distance of the torso for the repeatable gait cycle.

\subsubsection{Step1: moving legs}

For the update of footsteps, we can analytically solve the inverse-kinematics for the related three joint angles: $q_{4l-3}, q_{4l-2}$, and $q_{4l}$ as depicted in \figref{figure:design}.
Regarding the joint trajectory, following simple discrete path is proposed. First a certain angle is added for the first pitch joint $q_{4l-2}$ to lift the leg. Once the leg is lifted, the first yaw joint $q_{4l-3}$ and the second pitch joint $q_{4l}$ are moved to the target angles. Finally $q_{4l-2}$ is lowered to the target angle.
Given that there is no tactile sensor on the foot, thresholding method is used to determine the touchdown. Once all feet move to the new positions with a certain duration, all legs are considered to touch the ground. However, the new robot pose might not perfectly match the target pose. Thus, the target joint angles are finally reset to the current angles to clear unnecessary joint torques. During the movement of legs, the balance of the whole robot is guaranteed by the vectorable thrust proposed in \subsecref{crawling_control}, along with the decoupled joint torque control of \eqref{eq:joint_pd_control}.


\subsubsection{Step2: moving torso}

Once the legs have moved to the new positions, the torso will be lifted by all legs that are touching to the ground. The target position of torso is updated with the same gait stride length. I also design a simple discrete path for the torso movement, which starts from ascending with certain height, then moving horizontally, and finally descending to the new position.
Inverse-kinematics is used to calculate the joint angles for each key pose  according to the new torso position and the current foot position.
The torso movement is also achieved by using both the joint torque and the thrust control. The contact force is also considered to obtain the vectorable thrust control which is proposed in our previous work \cite{aerial-robot:SPIDAR-RAL2023}. Again, since there is no tactile sensor on the bottom of the torso, a similar thresholding method is applied to determine the touchdown. After the touchdown, the target joint angles are also reset to the current angles to clear the joint torque.

\section{Experiments}
\label{sec:experiment}

\subsection{Robot Platform}
In this work, I used the prototype of SPIDAR that was developed in our previous work \cite{aerial-robot:SPIDAR-RAL2023}.
This robot was composed of four legs and eight links as shown in \figref{figure:intro}.
The range of joint angle was $\left[-90^{\circ} \ 90^{\circ}\right]$. The maximum joint speed was set to  \SI{0.2}{rad/s} to satisfy the quasi-static assumption in the presented control framework. A hemisphere foot with anti-slip tape was equipped to ensure the stable point contact during the terrestrial locomotion.
For the vectorable rotor module, a pair of propellers were enclosed by ducts with the aim of safety and increase of thrust. This rotor module was allocated in the middle of each link. Eight Lipo Batteries (6s \SI{3000}{mAh}) were distributed in each link unit in parallel with the aim of weight distribution. 
At the center of the torso, NVIDIA Jetson TX2 was deployed to perform the realtime control framework as presented in \figref{figure:control}. The optimization problem of \eqref{eq:rough_allocation_cost} $\sim$ \eqref{eq:joint_constraint} was solved with OSQP solver, which led to a relatively fast computation time within \SI{5}{ms} and allowed the control loop to run at \SI{40}{Hz}. An external motion capture system was applied in our experiment to obtain the ego-motion state of the robot. More detail can be found in \tabref{table:specification}.

\begin{table}[h]
  \begin{center}
    \caption{Prototype Specifications}
    \vspace{-1mm}
    \begin{tabular}{c|ccc|ccc|c}
      \multicolumn{2}{c}{1. Main Feature} && \multicolumn{2}{c}{2. Vectorable Rotor}  && \multicolumn{2}{c}{3. Link and Joint}\\
      total mass & \SI{16}{kg} && rotor KV &  1550 && link diameter & \SI{0.025}{m} \\ \cline{1-2} \cline{4-5} \cline{7-8}
      max size (dia.) & \SI{2.7}{m} && propeller diameter & \SI{5}{inch} && link length & \SI{0.54}{m} \\ \cline{1-2} \cline{4-5} \cline{7-8}
      max flight time &  \SI{9}{min} && max thrust ($\bar{\lambda}$) & \SI{42}{N}  && max torque ($\bar{\tau}_q$)  & \SI{7}{Nm} \\ \cline{1-2} \cline{4-5} \cline{7-8}
      max crawl time &  \SI{25}{min} &&  max vectoring torque & \SI{1.5}{Nm} && max joint speed & \SI{0.2}{rad/s}\\ \cline{1-2} \cline{4-5} \cline{7-8}
    \end{tabular}
    \label{table:specification}
    \vspace{-5mm}
  \end{center}
\end{table}

\subsection{Flight}
\begin{figure}[!t]
  \begin{minipage}{\hsize}
  \begin{center}
    \includegraphics[width=0.95\columnwidth]{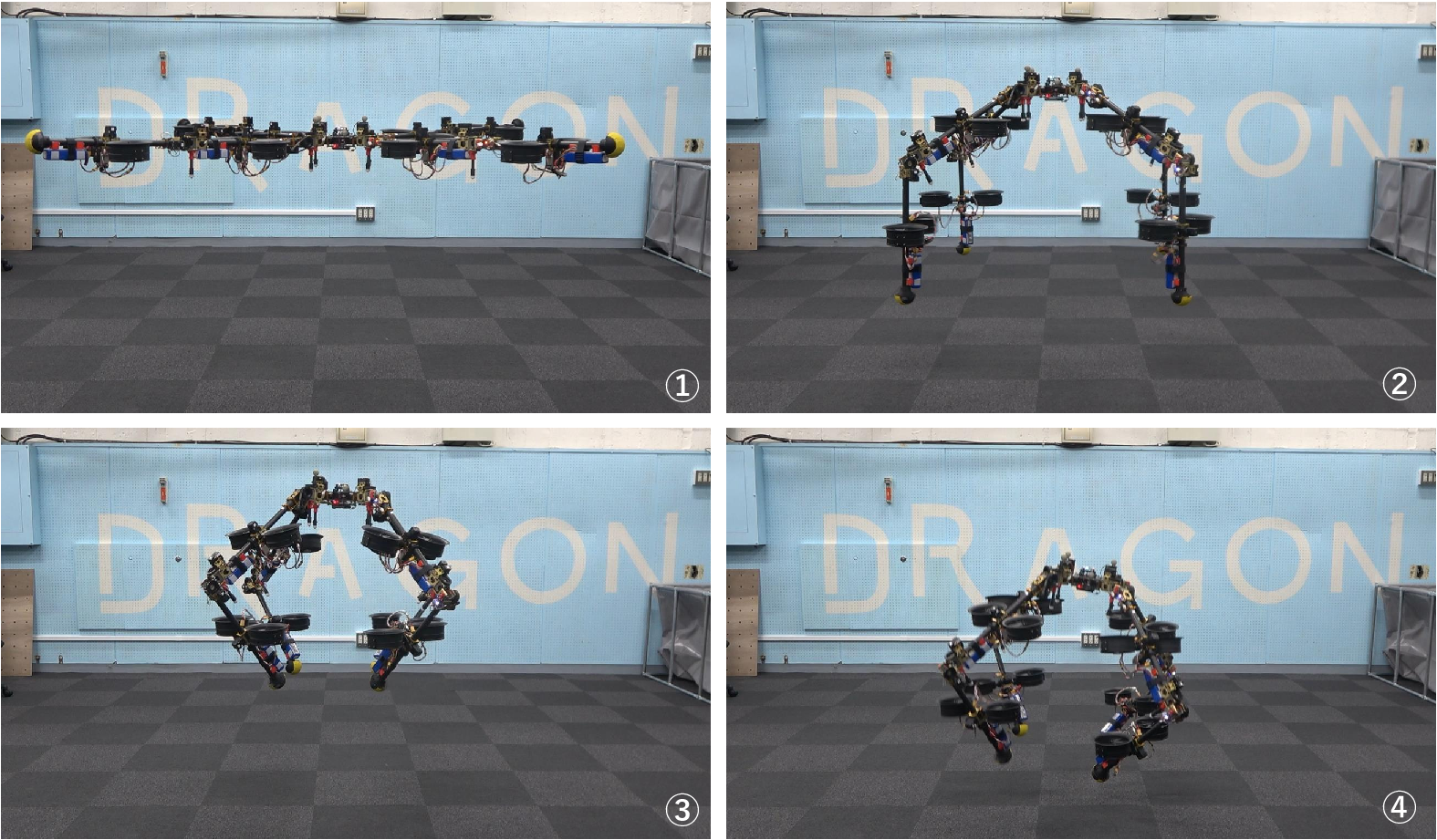}
    \vspace{-3mm}
    \caption{{\bf Different joint configuration while flying.} Form $\textcircled{\scriptsize 1}$: $q_i = 0$; form $\textcircled{\scriptsize 2}$: $q_{4l-3} = q_{4l-1} = 0, q_{4l-2} = 30^{\circ}, q_{4l} = 60^{\circ}$; form $\textcircled{\scriptsize 3}$: $q_{4l-3} = q_{4l-1} = 0, q_{4l-2} = 45^{\circ}, q_{4l} = 90^{\circ}.$ $l\in \{1, 2, 3, 4 \}$. Form $\textcircled{\scriptsize 4}$ was identical to form $\textcircled{\scriptsize 3}$, but without the constraint presented in \secref{flight}.}
    \label{figure:flight_expe_images}
  \end{center}
  \end{minipage}
  \begin{minipage}{\hsize}
    \begin{center}
      \vspace{3mm}
    \includegraphics[width=1.0\columnwidth]{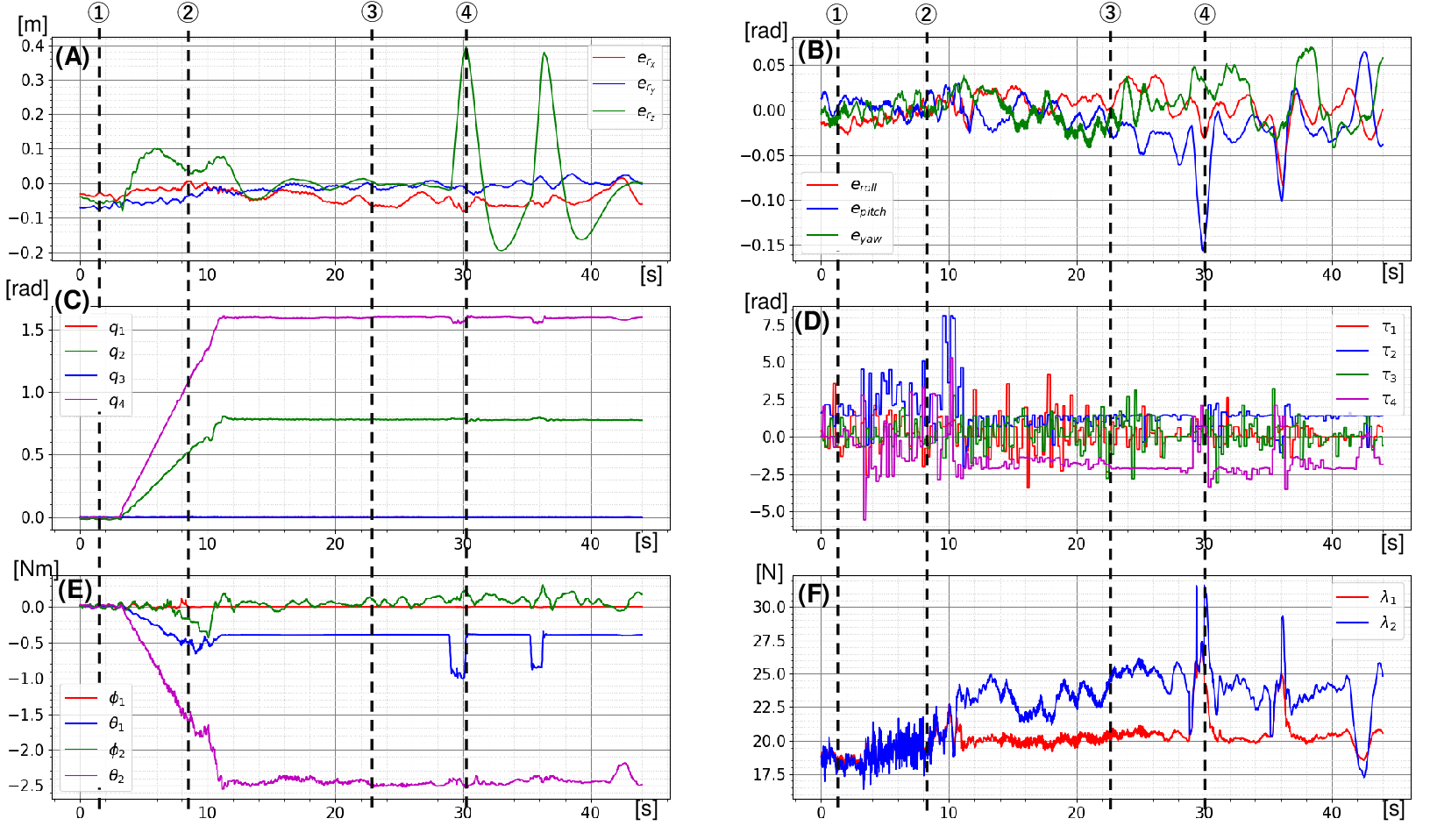}
    \vspace{-5mm}
    \caption{{\bf Plots related to \figref{figure:flight_expe_images}}. {\bf (A)/(B)}: centroidal positon and orientation (euler angles) errors. {\bf (C)/(D)}: trajectories of the angles and torques of joints in the first leg. {\bf (E)/(F)}: trajectories of the vectoring angles and the thrust force of the rotors in the first leg.}
    \vspace{-5mm}
    \label{figure:flight_expe_plots}
  \end{center}
  \end{minipage}
\end{figure}

To evaluate the performance of flight with joint motion, three typical forms were designed as shown in \figref{figure:flight_expe_images}. Form $\textcircled{\scriptsize 1}$ ($q_i = 0$) was a totally flat pose; whereas form $\textcircled{\scriptsize 2}$ ($q_{4l-3} = q_{4l-1} = 0, q_{4l-2} = 30^{\circ} , q_{4l} = 60^{\circ}$) contained four vertical links, leading to a singularity for the two-DoF vectoring apparatus. According to the planning method for the two-DoF vectoring apparatus proposed in \cite{aerial-robot:DRAGON-RAL2020}, it is possible to solve this singularity by locking  the corresponding vectoring angle $\phi_i$.
Another symmetric form $\textcircled{\scriptsize 3}$ ($q_{4l-3} = q_{4l-1} = 0, q_{4l-2} = 45^{\circ}, q_{4l} = 90^{\circ}$) was a more challenging configuration, because there were four pairs of rotors that overlapped with each other in the vertical direction. To solve the interrotor aerointerference, the extended thrust control method proposed in \subsecref{vectored_thrust_control} was adopted.


During \SI{0}{s} $\sim$ \SI{27}{s}, the centroidal position errors along each axis was almost within \SI{0.05}{m} as depicted in \figref{figure:flight_expe_plots}(A); however $e_{r_z}$ showed a large deviation during \SI{4}{s} $\sim$ \SI{12}{s}. I considered this was because the relatively fast joint motion broke the quasi-static assumption for the centroidal motion described in \subsecref{modeling}.
For the rotational motion, the errors were almost within \SI{0.05}{rad} as depicted in \figref{figure:flight_expe_plots}(B).
The overall RMSE for position were [0.038, 0.037, 0.043] \si{m}, and that of orientation were [0.016, 0.016, 0.016] \si{rad}, which demonstrated the desired stability and robustness of flight with joint motion.
During 11\si{s} $\sim$ 28\si{s}, the robot was under the form $\textcircled{\scriptsize 3}$. According to the proposed control method, the vectoring angles $\phi_1, \phi_3, \phi_5, \phi_7$ were fixed at zero, and the vectoring angles $\theta_1, \theta_3, \theta_5, \theta_7$ should be also titled to not point toward the lower rotors. Therefore these angles were almost fixed at \SI{-0.4}{rad} as depicted in \figref{figure:flight_expe_plots}(C), which is equal to the boundary $\underline{\mu}_{\hat{S}_{i}}$ in \eqref{eq:theta_bound}. Since the upper rotors were all titled, the thrust force of each rotor $\lambda_i$ was larger than $m_{\Sigma}/8$ (=\SI{20}{N}). This can be observed from \figref{figure:flight_expe_plots}(F), where the averages of both $\lambda_1$ and $\lambda_2$ became large than those under the forms of $\textcircled{\scriptsize 1}$ and $\textcircled{\scriptsize 2}$. For joint torque, the large load of $\tau_2 $ and $\tau_4$, which corresponded to the change in the joints  $q_2$ and $q_4$, can be observed during the joint motion (\SI{4}{s} $\sim$ \SI{11}{s}) as plotted in \figref{figure:flight_expe_plots}(D). When the robot was under a fix configuration, the joint load was relatively constant. The largest torque load under the fixed form was related to  $q_2$, of which the torque $\tau_2$ was around \SI{2}{Nm}; however, this value was much smaller than the limit $\bar{\tau}_q$.

To compare with the baseline method, the additional constraints \eqref{eq:inf_linear_bound} $\sim$ \eqref{eq:y_linear_bound} were temporarily removed from the control allocation during \SI{28.5}{s} $\sim$ \SI{30}{s} and \SI{35}{s} $\sim$ \SI{36.5}{s}. This resulted in an instant descending because of the interrotor aerointerference as shown in \figref{figure:flight_expe_images}$\textcircled{\scriptsize 4}$ and \figref{figure:flight_expe_plots}(A), which was quickly recovered after switching back to the extended thrust control method.

\vspace{-1mm}
\subsection{Crawling Motion}
\begin{figure}[!t]
  \begin{minipage}{\hsize}
    \begin{center}
      \includegraphics[width=0.95\columnwidth]{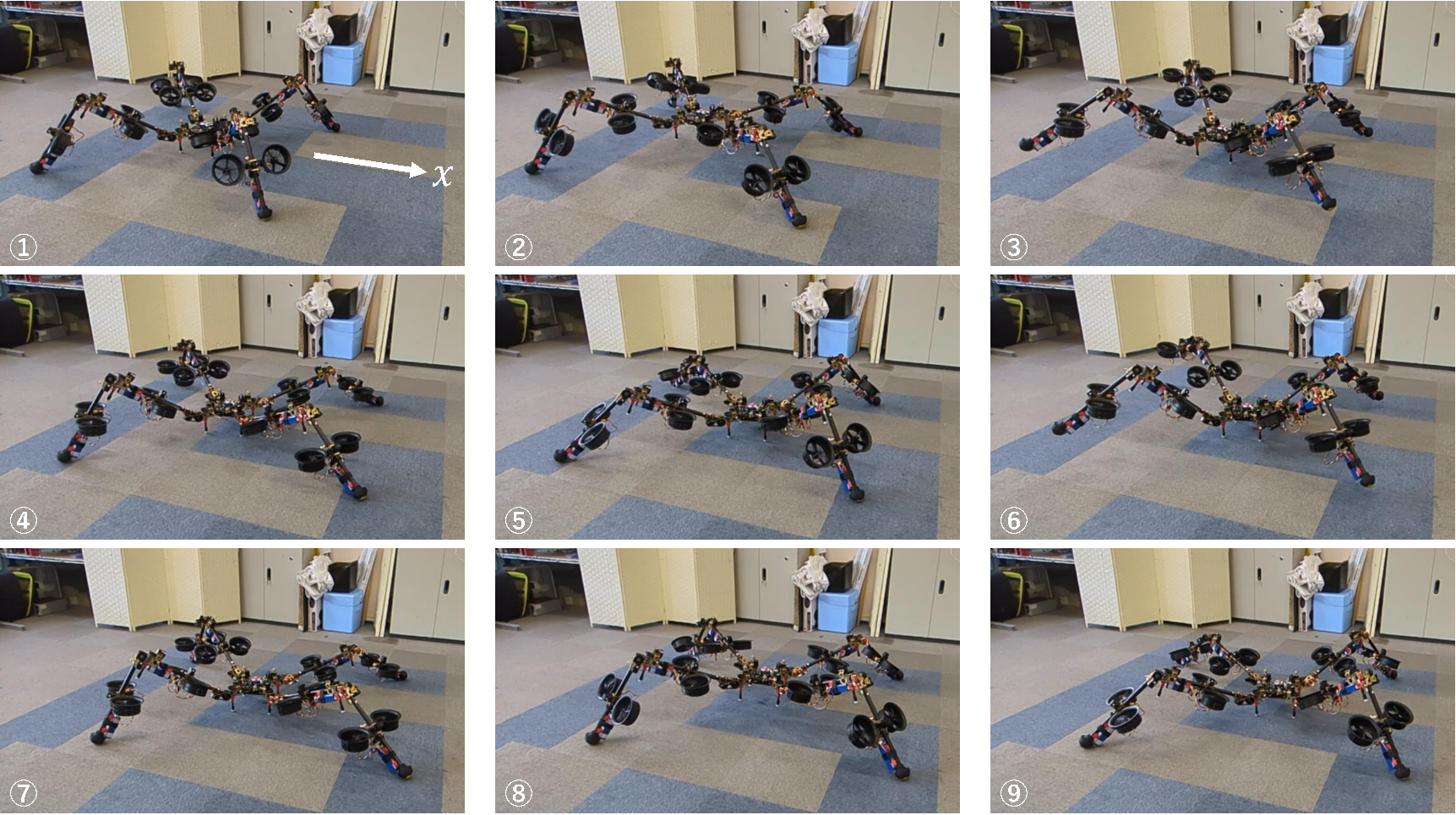}
      \vspace{-4mm}
      \caption{{\bf Crawling motion}. The robot performed 3 gait cycles to move \SI{0.6}{m} along the $x$ axis. I also evaluated the crawling motion with other directions which can be found in the supplementary video.}
      \label{figure:crawl_expe_images}
    \end{center}
  \end{minipage}
  \begin{minipage}{\hsize}
    \begin{center}
      \vspace{3mm}
      \includegraphics[width=1.0\columnwidth]{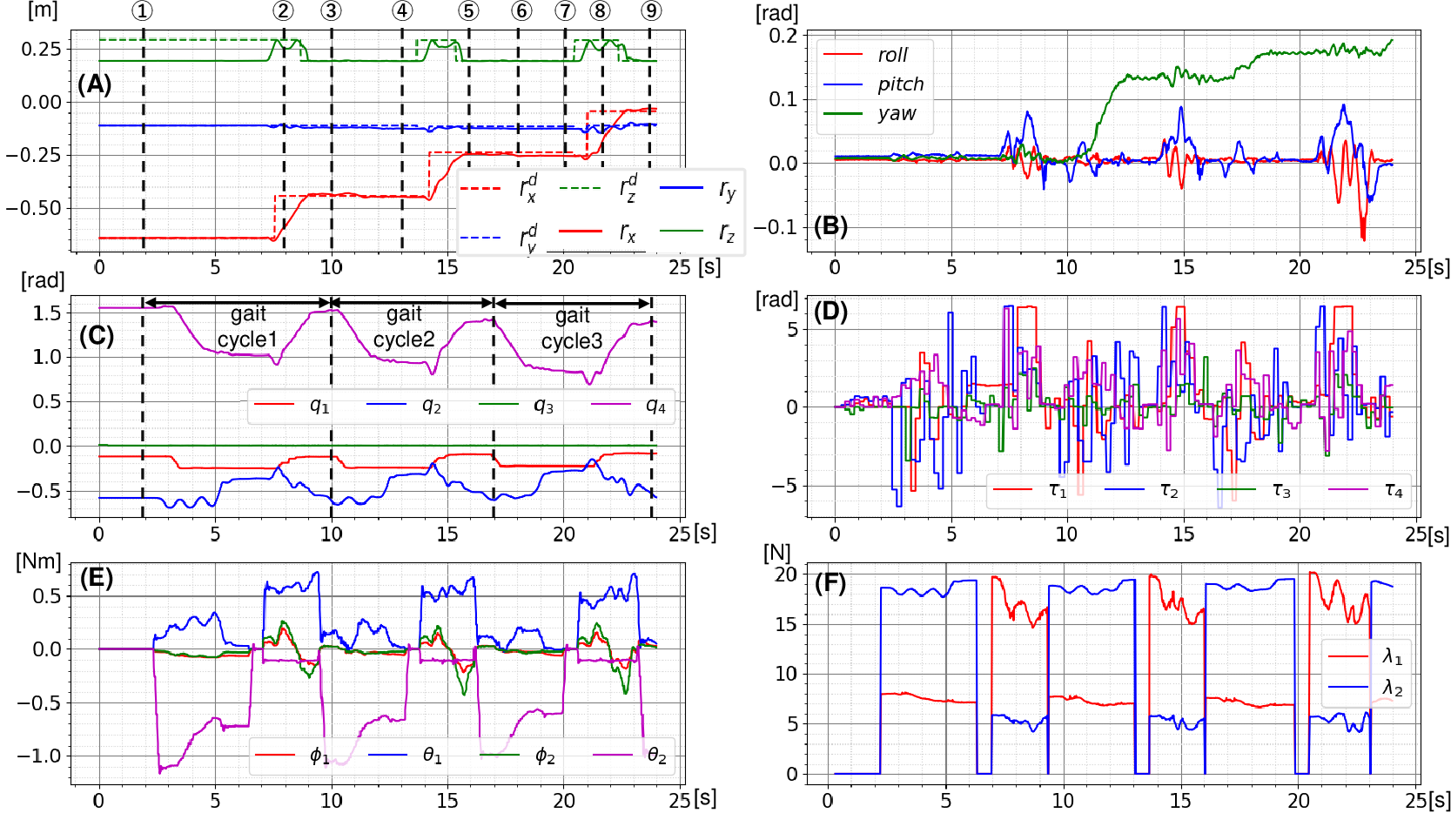}
      \vspace{-7mm}
      \caption{{\bf Plots related to \figref{figure:crawl_expe_images}}. {\bf (A)/(B)}: trajectory of the torso pose. {\bf (C)/(D)}: trajectories of the angles and torques of joints in the first leg. {\bf (E)/(F)}: trajectories of vectoring angles and the thrust force of the rotors in the first leg.}
      \label{figure:crawl_expe_plots}
    \end{center}
  \end{minipage}
\end{figure}

To evaluate the repeatable performance of crawling motion, I designed a forward movement of \SI{0.6}{m} along the $x$ axis on a flat ground as shown in \figref{figure:crawl_expe_images}. This movement required 3 gait cycles with a stride of \SI{0.2}{m}. Then the trajectory of joint angles, along with the thrust to lifting legs and torso, which are depicted in \figref{figure:crawl_expe_plots}, were calculated according to the method described in \secref{crawl}. Given that all legs have the similar trajectories, I only depict the states related to the first leg.

As shown in \figref{figure:crawl_expe_plots}(A), the torso was lifted \SI{0.1}{m} above the ground to move forward. After three gait cycles, the torso position reached the final goal, and the final position errors were [0.010, 0.010, 0.005] \si{m}, demonstrating a relatively accurate translational movement. Regarding the rotational motion of the torso, there was a relatively large drift around the yaw axis when all legs were lifted as shown in \figref{figure:crawl_expe_plots}(B). This can be attributed to the slightly different motion between legs (especially when the leg was swing horizontally) that caused the undesired moment around the yaw axis. Besides, the pitch angle of torso was also unstable when all leg was lifted (\SI{9}{s} $\sim$ \SI{11}{s} and \SI{16}{s} $\sim$ \SI{18}{s} in \figref{figure:crawl_expe_plots}(B)). This was because the relatively fast joint motion for lifting legs broke the quasi-static assumption for the proposed control method in \subsecref{crawling_control}, leading the unexpected additional moment on the torso. Moreover, the joint angles, especially $q_2$, were unstable when related leg was lifted as shown in \figref{figure:crawl_expe_plots}(C). This also corresponded to the fluctuation of joint torque $\tau_2$ as plotted in \figref{figure:crawl_expe_plots}(D).
I consider this was due to the lack of the joint stiffness that was provided by the weak servo motor. In our control framework, the thrust force was only used to compensate the gravity. Then, the feedback control by the thrust force based on the dynamics model of legs can be considered as an effective solution to solve the unstable issues in both the torso motion and leg motion.
In spite of the future work to improve the stability, the successful forward movement demonstrated the feasibility of the proposed method in terms of locomotion. 

As shown in \figref{figure:crawl_expe_plots}(F), the total thrust required for crawling was less than half of that required for flight, allowing significantly longer operation in terrestrial mode.
Currently, the robot required approximately \SI{20}{s} to traverse \SI{0.6}{m}, though the torso movement itself only took \SI{6}{s}. Therefore, it is possible to increase the velocity by making the leg lifting motion faster.


\vspace{-3mm}

\section{Conclusion}
\label{sec:conclusion}
In this paper, I presented the achievement of the unique aerial/terrestrial locomotion capability of quadruped robot SPIDAR equipped with the vectorable rotors. First, I identified interrotor aerointerference related to the rotor vectoring, and derived  explicit boundaries for the rotor vectoring range, which can be directly integrated into the flight control framework. I then developed the fundamental control method and the gait strategy for a special crawling motion that utilizes the vectorable thrust force to assist the weak joint actuator. The feasibility of the above methods were verified by the flight and crawling experiments with the prototype of SPIDAR.

A crucial issue remained in this work is the unstable motion of the legs and the torso during crawling. To improve the stability, the feedback control by thrust force based on  the joint dynamics will be developed to replace the current simple joint position control. Furthermore, the crawling motion would be performed on uneven terrain such as stairs to demonstrate its advantage in terrestrial locomotion.
Last but not least, the whole-body grasping in the unique configuration shown in \figref{figure:flight_expe_images}  $\textcircled{\scriptsize 3}$ will be explored to highlight the robot's versatility in both maneuvering and manipulation.

\bibliographystyle{styles/bibtex/splncs03_unsrt}
\bibliography{main}

\end{document}